\pdfoutput=1
\documentclass[11pt, table]{article}

\usepackage[final]{acl}

\usepackage{amsmath} % For mathematical symbols
\usepackage{algorithm}
\usepackage{algpseudocode}
\usepackage{amsfonts} % For mathbb if needed for other symbols, though not strictly for this
\usepackage{amssymb}
\newcommand{\simfunc}{\textbf{sim}}
\newcommand{\respred}{\textbf{res\_pred}}
\newcommand{\resactual}{\textbf{res\_actual}}
\newcommand{\finalscore}{\textbf{final\_score}\xspace}

\usepackage{times}
\usepackage{latexsym}
\usepackage{booktabs}
\usepackage{tcolorbox}
\usepackage{multicol}
\usepackage{cuted}
\usepackage{enumitem}
\setlist[itemize]{noitemsep, topsep=0pt}
\usepackage[T1]{fontenc}
\usepackage[utf8]{inputenc}

\usepackage{microtype}

\usepackage{inconsolata}

\usepackage{graphicx}

\newcommand{\tss}[1]{\textsuperscript{#1}}
\newcommand{\tn}[1]{\textnormal{#1}}
\usepackage{multirow}

\usepackage{subfigure}

\usepackage{amsmath} 

\usepackage{todonotes}

\usepackage{hyperref}
\usepackage{cleveref}

\newcommand{\eat}[1]{}

    \usepackage{colortbl} % For cell color
\usepackage{xcolor} % For defining colors
\usepackage{titlesec}
\titlespacing{\paragraph}
{0pc}{1.5ex minus .5 ex}{0.6pc}
\definecolor{pale-yellow}{HTML}{F5F176}
\newcommand{\highlight}[1]{\colorbox{pale-yellow}{#1}}

\definecolor{ocr}{HTML}{00C8FF}
\definecolor{ocr}{HTML}{009900}
\definecolor{jk}{rgb}{0.6, 0.2, 1.0}
\definecolor{jack}{rgb}{1.0, 0.6, 0.4}
\definecolor{owen}{rgb}{0.55, 0.71, 0.0}
\definecolor{pegah}{HTML}{F5F176}

\usepackage{xspace} 
\newcommand{\tsc}[1]{\textsc{#1}\xspace}
\newcommand{\gtv}{{Gram2vec}\xspace}
\newcommand{\rsp}{\tsc{RS}}
\newcommand{\rspfull}{{residualized similarity}\xspace}
\newcommand{\ic}{{\sc IntConf}\xspace}
\newcommand{\luar}{\tsc{LUAR}}
\newcommand{\elfen}{\tsc{ELFEN}}
\newcommand{\luarru}{$\textsc{LUAR}_{ru}$\xspace}

\title{Residualized Similarity for Faithfully Explainable Authorship Verification}

\author{
    Peter Zeng\tss{\tn{$\blacklozenge\spadesuit$}} 
    Pegah Alipoormolabashi\tss{\tn{$\blacklozenge$}}
    Jihu Mun\tss{\tn{$\blacklozenge$}} 
    Gourab Dey\tss{\tn{$\blacklozenge$}} 
    Nikita Soni\tss{\tn{$\blacklozenge$}} \\
    \bf{
     Niranjan Balasubramanian\tss{\tn{$\blacklozenge$}} 
    Owen Rambow\tss{\tn{$\clubsuit\spadesuit$}} 
    H. Andrew Schwartz\tss{\tn{$\blacklozenge$}} 
    }
    \\
    \tss{$\blacklozenge$}Department of Computer Science
    \tss{$\clubsuit$}Department of Linguistics\\
    \tss{$\spadesuit$}Institute for Advanced Computational Science\\
    Stony Brook University\\
    \texttt{pezeng@cs.stonybrook.edu}
}

\begin{document}
\maketitle
\begin{abstract}
Responsible use of \textit{authorship verification (AV)} systems requires not only high-accuracy but also \textit{interpretable} solutions. Specifically, for systems to be deployed in contexts where decisions have real-world consequences, their predictions must be explainable through interpretable features that can be traced to the original text. Neural methods achieve high accuracies, but their representations lack direct interpretability. Furthermore, LLM predictions cannot be explained faithfully -- if there is an explanation given for a prediction, it doesn't represent the reasoning process behind the model's prediction. To address this gap, we introduce \textit{\rspfull} (\rsp), \footnote{\url{https://github.com/peterzeng/rsp}} a novel method that supplements systems using interpretable features with a neural network to improve their performance while maintaining interpretability. Authorship verification is fundamentally a similarity task, where the goal is to measure how likely two documents are to be written by the same author. The key idea is to use a neural network to predict a \emph{residual similarity}, i.e. the error in the similarity predicted by the interpretable system. Our evaluation across four datasets shows that not only can we match the performance of state-of-the-art authorship verification models, but we can show how and to what degree the final prediction is faithful and interpretable.
% \pegah{I think this needs a small hint to why we want to predict "similarity". i.e. the connection between AV at the beginning of the paragraph and similarity at the end.}

\end{abstract}

\section{Introduction}

%Para 1: Introduce authorship verification as a task.  It is of utmost importance to forensic linguistics Describe the specific application of forensic linguistics. \cite{10.1093/llc/fqad061}, \cite{shuy1996language}, \cite{introtoforensic2016}  Establish need not only for explainable AI which is faithful, and also uses groundable evidence.  Need good vocabulary ("doubly faithful explainable AI"?  faithful and groundable explainable AI"? "FEGAI" for now -- faithfully explainable and groundable AI)

% Owen text
Identifying the author of a text or a collection of texts is a task with many use cases. In forensic investigations, stylometry techniques and authorship identification help link anonymous social media accounts \cite{Weerasinghe_Singh_Greenstadt_2022}, narrow down suspects \cite{10.1093/llc/fqad061}, and provide supporting evidence in court \cite{shuy1996language,introtoforensic2016}.  Plagiarism and academic dishonesty are cases of intentionally false authorship claims \cite{10.1145/3543895.3543928,10.1145/3292577}.  
%Andy: suggest new paragraph. Forensic and related uses...

\begin{figure}[t]
    \centering
    \includegraphics[width=\columnwidth]{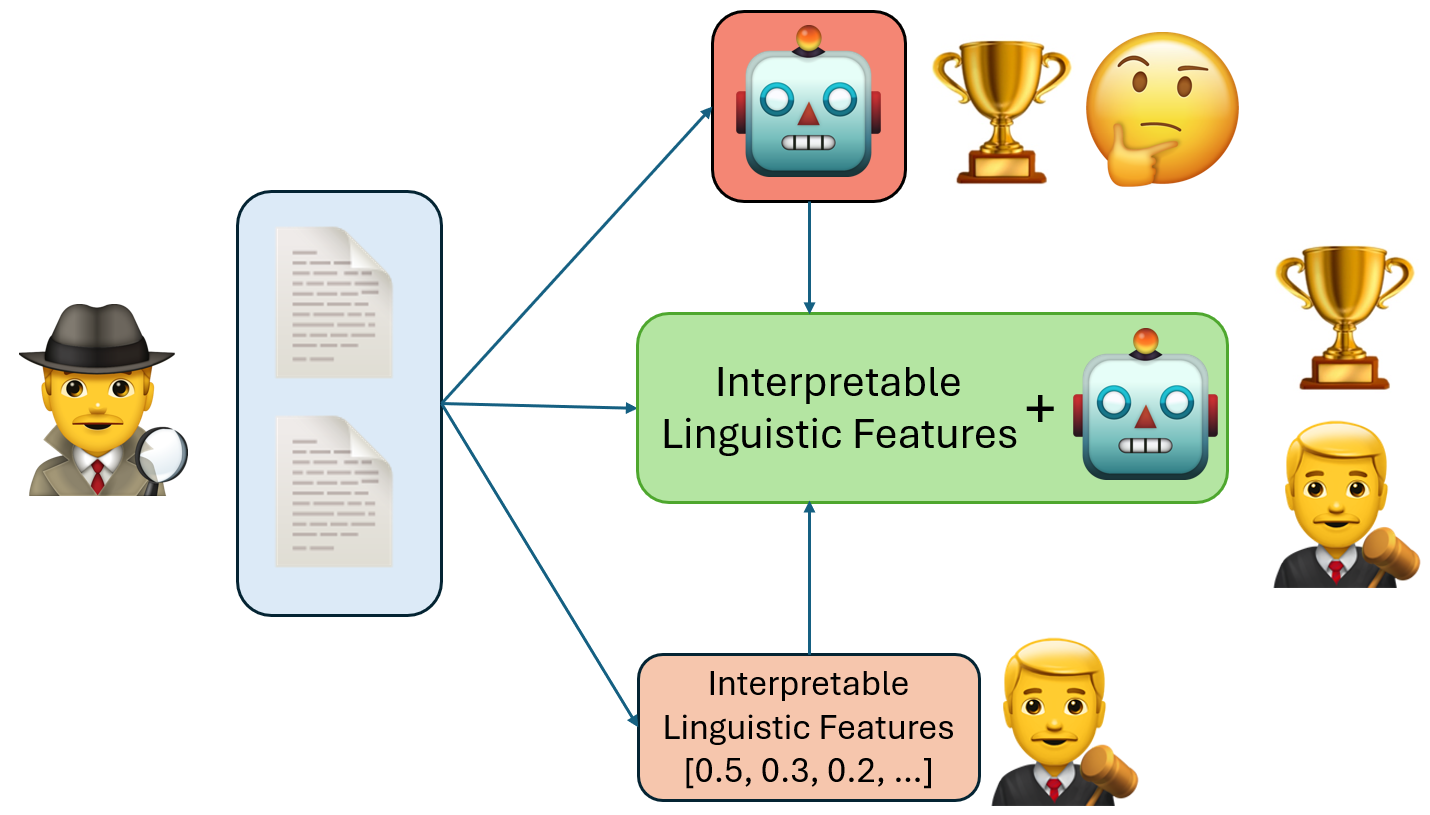}
    \caption{Demonstration of the task of Authorship Verification. A forensic linguist is trying to determine if two texts share the same author. They may use either an interpretable system comprising linguistic features faithful to the source text or a neural model, which has good performance but lacks interpretability. Our system combines the relative strengths of both by using a neural model to correct the error in the interpretable system’s prediction.}
    \vspace{-5pt}
    \label{fig:overview}
\end{figure}

Forensic and related applications of AI, and of authorship analysis models in particular, require explainable AI \citep{mersha2024explainable}. All stakeholders need to be able to verify the authorship claims made by any automated system. Furthermore, the explanation must be {\em faithful}: it ``should accurately represent the reasoning process behind the model’s prediction" \cite{lyu2024towards}.  But this is not enough: the reasoning process itself must be based on {\em interpretable} features derived from the analyzed texts, i.e., the features must be meaningful to humans. 
%If a text is represented by an uninterpretable embedding, which is then used in a reasoning process that can be explained to a human, then the human still cannot trust the system's authorship claims.
Authorship claims from a system using uninterpretable text embeddings cannot be trusted, even if they are used in an explainable reasoning process. 
But interpretable features are still not enough. Even if the input features are interpretable, they need to be {\em traceable}. This means that each feature can be traced back to the text(s) being analyzed, and the value of the feature is based on reproducible evidence from the text. A traceable feature 
%may be a low-level feature which can be derived from the input text, or a higher-level feature which is composed from low-level features.  It 
cannot be a feature on whose value reasonable observers may disagree after examining the text, such as ``formality".  

As with many NLP tasks, authorship systems that use representations derived from neural language models often achieve better verification performance than interpretable representations do \citep{devlin2018bert, vaswani2017attention}.
However, neural representations are limited in many critical domains because they are not directly interpretable. When attempts are made to interpret predictions, such as by \citet{alshomary2024latent}, the explanations for a model's predictions are not guaranteed to be faithful to how the prediction was made.
%
%In this work, we ask the following question:\textit{How can we combine the relative merits of two approaches?}
%
%Gram2Vec \citep{gram2vec2023} is an algorithm that generates a set of \textit{interpretable} linguistic features by extracting normalized relative frequencies of grammatical features present in a text. This vector representation captures complex patterns in an author's writing and offers a strong starting point for the task of authorship verification (AV).  
%
%Using cosine similarity between document embeddings has become a standard method of discerning authorship \citep{castro-castro-etal-2015-authorship}, and there exist many strongly performing document embeddings using neural models such as RoBERTa \citep{DBLP:journals/corr/abs-1907-11692} and SBERT \citep{reimers-gurevych-2019-sentence}. More recently, there have been neural models designed specifically to capture stylistic representations of an author's style such as LUAR \citep{rivera-soto-etal-2021-learning}, which achieved state-of-the-art results on AV in three domains: Amazon reviews, fanfiction stories, and Reddit comments. However, LUAR is a black box model\owen{Well, it isn't, we know how it was trained etc etc, and we can even inspect teh weights in it if we wanted to} whose representations (embeddings generated for documents)\owen{I added this phrase about representations} lack interpretability. Gram2Vec offers this inherent interpretability, at the cost of performance when compared to a model like LUAR.
%
In this paper, we ask how one can combine the relative strengths of the two methods: the interpretability and traceability of linguistic representations and the high performance of neural models. 

% One way of doing so is direct ensembling, which involves determining fixed weights that combine scores from both an interpretable system (i.e., a system which uses only interpretable representations)  and a neural system. However, when this ensembling method is optimized for performance, the weight of the interpretable system is small, and when the the method is optimized for interpretability, the performance decreases. What we want, instead, is a more dynamic approach, one where we can rely on the interpretable system more when it is likely to be accurate, and rely on the neural model otherwise.

As the main contribution of the paper, we introduce \emph{residualized similarity} (\rsp), which uses the idea of estimating the \emph{residual} of a predictor i.e., the error in a model's prediction. We approach Authorship Verification as a similarity task, as is standard in the field. For each pair of documents we obtain some similarity score from a system; if the score is above a certain threshold, we conclude the documents share the same author, and if the score is below the threshold, we say the documents are from different authors.
% \niranjan{Andy: I am not sure if this is a correct description of the term residual in a formal statistical sense. If there are any differences it might be worth adding a note here.}
Suppose we start with an explainable system using interpretable and traceable features
%\owen{We could replace this by "with a FETA system"} 
as the initial similarity estimator. We can then train a neural model as a \emph{residual predictor}, which predicts the error or correction to the interpretable system's similarity score. The final prediction is a simple sum of the interpretable model's similarity score and the predicted residual, i.e., a similarity adjustment made by the neural model. 

This combined system can achieve the trade-off we desire: (i) when the interpretable model is likely to be correct, the residual should be low, providing interpretability and faithfulness while remaining accurate, and (ii) when the interpretable model is likely to be incorrect, the residual should provide the necessary correction, improving accuracy but reducing interpretability to a degree proportional to the error. 
This approach is inspired by the \textit{residualized control} approach~\citep{zamani-etal-2018-residualized}, which trains a residual model for a regression problem, combining numerous linguistic features with a few interpretable health-relevant attributes to predict community health indicators.  
We describe our approach in detail in Section~\ref{sec:rsp}.

%One such interpretable feature system, 
We use \gtv~\citep{gram2vec2023} as our interpretable feature system, which records normalized frequencies of morphological and syntactic features for input texts.  
We evaluate our \rsp approach by combining \gtv with a state-of-the-art AV neural model, \luar \citep{rivera-soto-etal-2021-learning}, finding that \rsp can match the performance of using \luar alone, while introducing faithful explanations using interpretable and traceable features. 
%We make a distinction between two aspects of faithfulness. First, our system's prediction can be explained directly using the underlying features in \gtv. Second, these features are directly measurable within a text, i.e. we can explain exactly why a feature in a given text has a certain value. 
We perform a case study on how our system retains interpretability, measured by an \textit{interpretability confidence (IC)} metric, which indicates the extent to which the interpretable system is used for a given input. Details of this are in Section \ref{sec:example}.

\section{Related Work}
\label{sec:rel}
Authorship verification, authorship attribution, and authorship profiling are all part of authorship analysis, which has been explored through a wide range of approaches (see surveys~\citet{el2014authorship, misini2022survey, 10.1145/3715073.3715076}). 
% \pegah{Here is a more recent survey if you want to cite it \cite{10.1145/3715073.3715076}}
% Here we discuss interpretable methods that make use of stylometric features and recent neural models.

\paragraph{Interpretable Methods} Previous stylometric approaches \citep{Stamatatos2016AuthorshipVA} often make use of readily interpretable features to train classifiers. Some examples include lexical features such as vocabulary, lexical patterns \cite{mendenhall-1887, van-halteren-2004-linguistic}, syntactic rules \citep{varela2016computational}, and others. \gtv, the interpretable component of \rspfull falls into this category.
% \owen{Should we add here that our system with interpretable features, g2v, falls into this paradigm as well -- i.e., we do not claim research novelty for it} 

\paragraph{Neural Models} Authorship verification has benefited from models built upon RNNs \citet{gupta-authorship}, CNNs \cite{hossain2021authorship}, BERT-like architectures \cite{manolache2021transferring}, and Longformers \cite{ordonez2020will, nguyen-etal-2023-improving}. More recently, sentence-transformer-based models \cite{wegmann-etal-2022-author, rivera-soto-etal-2021-learning} have obtained state-of-the-art performance for AV tasks. As we are interested in improving the performance of interpretable authorship verification, we focus on these SOTA AV models, particularly \luar \citep{rivera-soto-etal-2021-learning}. 

% have successfully applied recurrent neural networks (RNNs), long short-term memory networks (LSTMs), and gated recurrent units (GRUs), demonstrating significant improvements in authorship attribution.  TODO: add more % LUAR integrates an attention-based architecture with a contrastive training objective, setting new state-of-the-art (SOTA) benchmarks for AV across three distinct domains. \citet{wegmann-etal-2022-author} introduced a new Contrastive Authorship Verification (CAV) framework to control for content, employing Siamese BERT-based networks, enabling the extraction of more content-independent stylistic representations.

Our work uses residual similarity analysis to combine interpretability and neural models' high performance for authorship verification. Similar residual approaches have been used previously for improving performance in health outcome prediction, by combining lexical and health-relevant attributes~\cite{zamani-etal-2018-residualized}, and in a recent work that combines statistical and neural methods for machine translation~\cite{benko2024use}. Other works have focused on generating explanations, often layering other mechanisms on top of interpretable input features~\cite{boenninghoff-explainable,10.1145/3654675,9746262} or doing a post-hoc evaluation on a latent, non-interpretable space \cite{alshomary2024latent}. Some recent work also explores prompting large language models to derive interpretable stylometric features for authorship analysis~\cite{hung2023wrote,patel-etal-2023-learning}. However, these features are not 
traceable
%measurable 
in a text as the approaches rely on LLMs to generate the features, and the generations do not represent the reasoning process behind attributing a set of features to a text.
\section{Residualized Similarity}
\label{sec:rsp}

% \nb{I am writing a paragraph here that clearly states the problem that the residualized similarity is solving. Feel free to edit/delete for space or any other reason.}

\subsection{Problem Statement:} As we argued earlier, we want to develop an authorship verification system for applications such as forensic linguistics, where we want the system to use \textit{traceable interpretable} features as much as possible but also be highly accurate. To this end, we introduce an interpretable variant of the authorship verification problem, which not only requires an output decision but also requires an explicit confidence metric that shows the extent to which the decision can be attributed to the traceable interpretable features. On the one extreme, linguistic feature-based classifiers (e.g., Gram2Vec~\cite{gram2vec2023}) will have an interpretability confidence that is $100\%$ but lower accuracy, while on the other extreme, a neural embedding-based classifier~\cite{rivera-soto-etal-2021-learning} will likely have higher accuracy but $0\%$ interpretability confidence. The goal then is to design a system that achieves high accuracy and high interpretability confidence."

\begin{figure}[t]
    \centering
    \includegraphics[width=\columnwidth]{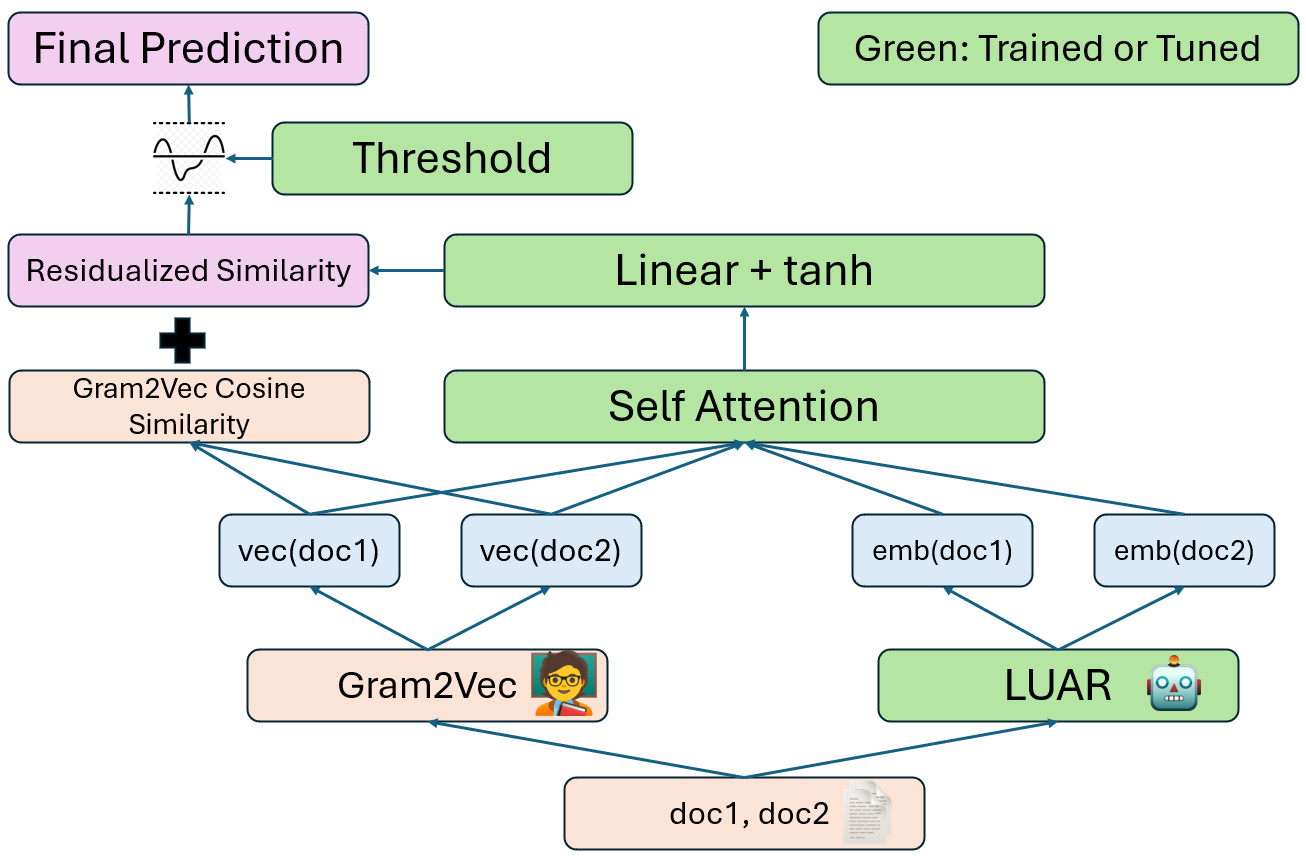}
    \caption{\textbf{Residualized Similarity Architecture.} 
   To incorporate signal from the interpretable feature vectors, we add an attention layer over both the interpretable feature vectors as well as the neural embeddings from the model we're fine-tuning. Boxes colored in green indicate that they're updated during training. On the left-hand side, we show the system in use at inference time. The final similarity score is a simple sum of the interpretable cosine similarity score and the predicted residual.}
   % We add a sequential layer, alternating linear and ReLU layers onto the encoder model to output the regression value, which we then pass through a tanh activation function. This is done to introduce non-linearity and capture more rich information in our fine-tuning.}
    \vspace{-5pt}
    \label{fig:architecture1}
\end{figure}
% \subsection{Residualized Similarity}
%%% Pegah's rewrite of this paragraph:
\subsection{Method Description}
To address the above-mentioned problem, we use \rspfull (\rsp), whose key idea is to train a neural model to predict the residual similarity, i.e., the difference between the cosine similarity obtained from the interpretable system and the ground truth. Per each train/dev/test set, we first generate interpretable feature vectors for each document using \gtv. Next, to account for difference in variance, the feature vectors are standardized (z-scored) per feature against their respective dataset. Finally, the cosine similarity is calculated between the standardized pairs of vectorized documents. The ground truth label is 1 for a pair of documents written by the same author and -1 otherwise. \rsp is trained to predict a residual ${r}^{(i)} = y^{(i)} - \textbf{sim}(f(d_1), f(d_2))$, where $y$ is the gold label, 
% \nb{here we are using $y$ to denote gold label (-1 or 1) but later in training objective $y^{(i)}$ refers to residual. My suggestion is you use $y^{(i)}$ to denote the gold label, as before but use $r^{(i)}$ for ground-truth residual and the \\hat versions of these for the corresponding predicted versions}
\textbf{sim} represents the cosine similarity between the pair of vectorized documents, $d_1$ and $d_2$ are the two documents, and $f$ is the \gtv vector function. We will call this the \textit{ground truth residual}. 

\Cref{fig:architecture1} illustrates the specifics of training the \rsp model and usage at inference time. The process of training \rsp begins with pairs of documents. These are vectorized both by the interpretable system and by the neural model we are fine-tuning, giving us four embeddings. Next, an attention layer is placed over all four embeddings, for \rsp to learn how much to weigh the interpretable features and the neural embeddings when making the residual prediction. Note that this step is only for the training stage. At inference time, given a pair of documents, the final similarity score is a simple summation of the cosine similarity from our interpretable system and the predicted residual. 

\paragraph{Training Objective}
Given a batch of document pairs $\mathcal{B} = \{(d_1^{(i)}, d_2^{(i)}, r^{(i)})\}_{i=1}^n$, where $r^{(i)}$ is the ground-truth residual for the pair $(d_1^{(i)}, d_2^{(i)})$, the model aims to predict the residual similarity, denoted as $\hat{r}^{(i)}$. The training objective is to minimize the Mean Squared Error (MSE) between the predicted and actual residuals:

$$
\mathcal{L}(\mathcal{B}) = \frac{1}{n} \sum_{i=1}^n \left(\hat{r}^{(i)} - r^{(i)}\right)^2
$$

%\nb{All of these terms are described already in the paragraph.}
%where, $n$ is the batch size
%\begin{itemize}
%    \item $n$ is the batch size,
%    \item $\hat{r}^{(i)}$ is the model's prediction for the residual,
%    \item $r^{(i)}$ is the ground-truth residual.
%\end{itemize}
% \nb{I was briefly confused about interpretability here. It would help to remind the reader here that this is only for training and during inference interpretability remains quite simple for the interpretable features during inference.}

%on all three English datasets.
% \nb{It is not clear what the contrastive-loss fine-tuned neural model is. Is it the current architecture you have in Figure 1?}

% Our evaluation tests how the \rspfull method fares against the performance of the two methods it combines: a system using only interpretable features, and a neural model fine-tuned on the target datasets. 

% For the neural model baseline, the neural model is trained on each dataset using a contrastive learning objective \citep{khosla2020supervised}.\owen{This is in 4.1 -- completely omit here?} 
%We evaluate the systems' performance based on the receiver-operating characteristic area under curve (AUC), as it is a way to measure performance of models that is threshold independent.\owen{This detail abot eval should go in section 5 and is already there} 

\subsection{Interpretability Confidence}
\label{sec:interpretability}
% We have shown that \rspfull is a hybrid system that uses a neural model to correct the error in prediction made by an interpretable system. In doing so, we can match the performance of a solely neural system, while retaining interpretability.
% In this section we discuss how to quantify the amount of interpretability a specific result retains.
\newcommand{\simgtv}{$\textbf{sim}_{\rm g2v}$}
\newcommand{\simgtvmath}{\textbf{sim}_{\rm g2v}}

We introduce the notion of ~\textit{``interpretability confidence''}(\ic), which is a way to measure how interpretable a particular prediction using \rspfull is.
We define \ic based on the final prediction. Let the similarity between two texts determined by \gtv alone be
\simgtv{}.
%i.e., a measure of the similarity between two texts. 
%put in relation 
Thus, \ic represents the relative contribution of \gtv to the overall similarity score,
%\nb{This clause is difficult to understand. Is it broken in some way?}, 
composed of the \gtv-derived score and the residual predicted by \rsp.  We distinguish two cases: the \rsp system predicts that the texts are by the same author, or it predicts different authors.  If the prediction is same author, the correct prediction is a high similarity score.  The contribution of \gtv is the distance from -1, the lowest similarity score possible, so we quantify the contribution of \gtv as $1 + \simgtvmath{}$, and divide it by 
%\nb{Maybe it is my ESL limitation but I dont understand this phrase ``put in relation''} 
the sum of the \gtv contribution and the contribution of the residual component, $|\text{predicted residual}|$.  Thus, we get: 
%If the prediction of the overall system is same author, then \ic is defined as 
$$\frac{1 + \simgtvmath{}}{1+\simgtvmath{} + |\text{predicted residual}|}$$ (or 0, if $1+\simgtvmath{} + |\text{predicted residual}| = 0$). If the prediction of the system is different author, then the contribution of the \gtv component is the distance from +1, the highest similarity score.  Thus, in this case, \ic is defined as: $$\frac{1 - \simgtvmath{}}{1 - \simgtvmath{} + |\text{predicted residual}|}$$  The \ic always takes values between 0 and 1.
% as have two parts, a score, defined as $1-|\text{predicted residual}|$, and an indicator of whether or not the label was flipped by the predicted residual (1 if flipped, 0 if not). The label is considered flipped if the cosine similarity prediction using \gtv is on one side of the cosine similarity threshold (different author if below, and same author if above), and adding the predicted residual from \rsp causes the final score to be on the other side of the threshold. We provide an example of this in Section~\ref{sec:example_2}. 
Note that we can calculate the \ic for any {\em specific} pair of documents after running \rsp. 

We note that sometimes the predicted residual ``flips" the prediction of the \gtv-only system, from ``same author" to ``different authors" or {\em v.v.}  This possibility is of course precisely why we are predicting the residual.  
We emphasize that even in cases where the prediction is flipped after using \rsp  we can still make use of the underlying interpretable system for interpretation. We show in section \ref{sec:example} that when the prediction was changed, the underlying interpretable system can help explain why a prediction was made.

% two methods it combines: an interpretable system,  neural models fine-tuned on the target datasets, as well as a weighted ensemble of the two. 

\section{Experimental Setup}
\label{sec:setup}

\looseness=-1 

We perform experiments across a variety of neural models, in order to evaluate robustness and ability to generalize of our technique. We specifically leverage sentence-transformer-based models \cite{reimers-gurevych-2019-sentence}, which were developed as a distinct technology in parallel to generative LLMs. Embedding models specialize in creating fixed-length semantic representations optimized for similarity computations--the way authorship verification is evaluated. 
\subsection{Models}
We evaluate a diverse set of pre-trained encoders to ensure the method is model-agnostic and robust across architectures. We include RoBERTa \cite{liu2019roberta} as a general purpose transformer and Longformer \cite{beltagy2020longformer} to handle long documents.
%beyond typical context lengths. 
We try specialized authorship representation models--LUAR \cite{rivera-soto-etal-2021-learning} and a Style-Embedding model, \cite{wegmann-etal-2022-author}--both pre-trained to capture distinctive style features. We also test all-mpnet-base-v2 \cite{song2020mpnet}, a sentence-transformer model at the top of the SBERT.net leaderboard, and a more recent model, mxbai-embed-large-v1 \cite{li2023angle, emb2024mxbai}, which achieves SOTA for BERT-large sized models on the Massive text embedding benchmark (MTEB) \cite{muennighoff-etal-2023-mteb}.

\subsection{Baselines} We compare \rspfull against two classes of baselines: the interpretable \gtv baseline, and the non-interpretable neural baselines. The models used for the neural baselines are mirrored in the models we tune in our system to predict the residual, all of which have less than one billion parameters. We fine-tune these models in a Siamese network using a contrastive loss function as the training objective. This approach is similar to SBERT \cite{reimers-gurevych-2019-sentence}, but we use the architecture to learn document-level, as opposed to sentence-level, embeddings.

% We highlight results using LUAR as it is state-of-the-art in the task of authorship verification  \citep{rivera-soto-etal-2021-learning}
% 
% We focus on LUAR and LUAR-RU, used for English and Russian text respectively. We choose LUAR as it is state-of-the-art in the task of authorship verification, and we also use the Russian version to demonstrate effectiveness across multiple languages. We fine-tune these models in a Siamese network using a contrastive loss function as the training objective. 

\subsection{Residualized Similarity Framework}
Let $d_1, d_2$ be two documents.
Let $y \in \{1, -1\}$ be the gold label, where $1$ indicates the same author and $-1$ indicates different authors.
Let $f$ represent the \gtv~vectorizer.
Let $\simfunc(v_1, v_2)$ be the cosine similarity function between two vectors $v_1$ and $v_2$.
The interpretable system's similarity score is $s = \simfunc(f(d_1), f(d_2))$.
The ground truth residual (actual residual) is defined as $\resactual = y - s$.
Let $\respred$ be the residual predicted by a neural model $M_{res}$. The model $M_{res}$ is trained to approximate $\resactual$.
The final similarity score is given by $\finalscore = s + \respred$.
Let $t$ be the classification threshold, which depends on preferences for false positives over false negatives and can be obtained by tuning on a held-out set.
%, such that the similarity must be more indicative of authorship than not. 
The final authorship prediction is obtained by comparing \finalscore  to the threshold: same author if greater than $t$, different author if less than $t$.

% The model parameters $\theta$ are updated using gradient-based optimization

% \noindent\textbf{Inference}:
% \begin{itemize}
%     \item For a new document pair:
%     \begin{itemize}
%         \item If $\textbf{final\_score} > t$: Predict same author
%         \item Otherwise: Predict different author
%     \end{itemize}
% \end{itemize}
\begin{table*}[ht]
\addtolength{\tabcolsep}{-0.2em}
    \centering
    \label{tab:neural_residual_performance}
    \begin{tabular}{lccccccccc}
        \toprule
         & \multicolumn{2}{c}{\textbf{Reddit}$_{AUC}$} & 
        \multicolumn{2}{c}{\textbf{Amazon}$_{AUC}$} & 
        \multicolumn{2}{c}{\textbf{Fanfiction}$_{AUC}$} &
        \multicolumn{2}{c}
        {\textbf{Pikabu}$_{AUC}$}\\
        \midrule
        % & Neural & Residual & Neural & Residual & Neural & Residual \\
        \textbf{G2V (Gram2Vec)} & \multicolumn{2}{c}{0.63} & \multicolumn{2}{c}{0.71} & \multicolumn{2}{c}{0.69} & \multicolumn{2}{c}{0.65} \\
        \midrule
        % \cmidrule(lr){2-3} \cmidrule(lr){4-5} \cmidrule(lr){6-7}
        \textbf{Neural Models} & Neural & \rsp & Neural & \rsp & Neural & \rsp & Neural & \rsp\\
        \midrule
        RoBERTa-Base & 0.69 & 0.72 & 0.87 & 0.85 & 0.88 & 0.87 & - & - \\
        RoBERTa-Large & 0.71 & 0.70 & 0.90 & 0.86 & \textbf{0.91} & 0.86 & - & - \\
        Longformer & 0.74 & 0.71 & 0.88 & 0.85 & 0.89 & 0.86 & - & -\\
       \luar/LUAR-RU & \textbf{0.84} & \textbf{0.80} & \textbf{0.91} & \textbf{0.90} & 0.89 & 0.87 & \textbf{0.74} & \textbf{0.76} \\
        Style & 0.77 & 0.76 & 0.83 & 0.83 & 0.72 & 0.78 & - & -\\
        all-mpnet-base-v2 & 0.66 & 0.62 & 0.86 & 0.84 & 0.90 & 0.85 & - & -\\
        mxbai-embed-large & 0.72 & 0.69 & 0.88 & 0.85 & 0.87 & \textbf{0.89} & - & - \\
        \bottomrule
    \end{tabular}
     \caption{Model performance across three datasets: Reddit, Amazon, and Fanfiction. \gtv represents the fully interpretable baseline, the neural columns represents the non-interpretable contrastive-loss fine-tuned baseline, and Residual is our system. Pikabu is a Russian dataset, and we use LUAR-RU as the neural model in \rsp (its English counterpart generally performed the strongest in the English datasets). The best performing system in each column is bolded.}
     \label{tab:full-results}
\end{table*}

\paragraph{Training Details} All neural models and \rsp are trained using LoRA~\citep{hu2021lora}, which reduces the number of trainable parameters and memory requirements. We observe that using LoRA also yields better performance overall for all models as compared to a full fine-tuning. For evaluation of system performance, we use receiver-operating curve area under curve (AUC), which doesn't require tuning of a threshold. Additional training details are in Appendix~\ref{sec:appendix_training}. 
\looseness=-1
% For our case study in section \ref{sec:example}, we select a threshold of 0.5 as a natural midpoint from 0 to 1, suggesting that documents need to be more similar than dissimilar to be considered by the same author.

\subsection{Data}
% \pegah{Maybe we can show the datasets in a table, and put the additional details in the appendix.}
% \nb{Can you move this to later in the draft? Why do we have experimental settings before method description? I'd say swap this with section 4.}
We train and evaluate our \rspfull system on four datasets
% \owen{We shold probably say we will share them or share scripts that create them} 
covering diverse genres. We choose the first three as they are the datasets used by \citet{rivera-soto-etal-2021-learning} from the original training of \luar, and we include the Russian dataset Pikabu to evaluate our method on another language as we had access to a Russian version of \luar.

In order to train both \rsp and the contrastive-loss fine-tuned baseline, we require the data to be in a labeled paired format: \{Document 1, Document 2\} and same/different author. 
% \pegah{If we have (doc1, doc2, label) then it is not a pair, right? It's a triplet. Or we can say labeled pairs?}
% The full details of training \rsp are provided in section \ref{sec:rsp}.
For the contrastive-loss fine-tuned baseline, the aim is to push pairs of documents by the same author together, and to push pairs of documents by different authors apart.

\paragraph{Reddit Comments} We use a dataset of Reddit comments from 100 active subreddits curated by ConvoKit \citep{chang-etal-2020-convokit}. We use a version preprocessed by \citep{wegmann-etal-2022-author}, as invalid comments, comments containing only some sort of white space, and deleted comments are removed.  We create pairs of comments, label them for author verification. Reddit comments can be naturally very short, so we further filter the comment pairs and keep only comments longer than 20 words. This results in roughly 50,000 training pairs, 10,000 validation pairs, and 10,000 testing pairs. For the rest of the datasets, in order to have comparable train/validation/test sizes, we randomly sample them to match the size of the Reddit train/validation/test splits.

\paragraph{Amazon Reviews} From the Amazon review dataset \cite{ni-etal-2019-justifying}, we take reviews from three categories: Office Products; Patio, Lawn and Garden; and Video games. We use a reduced dataset where all items and users have at least 5 reviews, and we keep authors with at least two reviews of 20 or more words. The validation set is split from the training set by taking stories from 1/6 of the authors. Then, we sample same author pairs by randomly choosing an author and two texts written by them. For different author pairs, two authors and one text from each author are randomly chosen.

\paragraph{Fanfiction Stories} The fanfiction dataset contains 75,806 stories from 52,601 authors in the training set and 20,695 stories from 14,311 authors in the evaluation set. We use the pre-processing script from \luar \citep{rivera-soto-etal-2021-learning} to split each story into paragraphs since fanfictions can be very long. The process of sampling pairs of reviews is the same as in the Amazon dataset. 

\paragraph{Pikabu comments} We start with the Pikabu dataset from \citet{ilya_gusev_2024} available on HuggingFace. We drop documents with fewer than 100 characters, and authors with fewer than two documents; we then anonymize the data, redacting credit card numbers, IP addresses, names, and phone numbers.

% \begin{figure}[h]
%     \centering
%     % \subfigure[]{\includegraphics[width=0.45\textwidth]{AUC-curves/reddit_auc.png}}
%     % \hspace{0.05\textwidth}
%     % \subfigure[]{\includegraphics[width=0.45\textwidth]{AUC-curves/amazon_auc.png}}
%     % \subfigure[]{\includegraphics[width=0.45\textwidth]{AUC-curves/fanfiction_auc.png}}
%     % \hspace{0.05\textwidth}
%     % \subfigure[]{\includegraphics[width=0.45\textwidth]{AUC-curves/pikabu_auc.png}}
%     \includegraphics[width=\columnwidth]{AUC-curves/amazon_auc.png}
%     \caption{ROC AUC Curves for \gtv, LUAR, and \rsp on the (a) Reddit, (b) Amazon, (c) Fanfiction, and (d) Pikabu datasets. We observe performance increases comparing \rsp to \gtv that range from 11 points on Pikabu to 25 points on Amazon. Notably, \rsp also sees a 2 point increase in performance from LUAR on Pikabu.}
%     \label{fig:auc}
% \end{figure}

For all four datasets, we use 50K, 10K, and 10K pairs for the training, validation, and test sets respectively. The ratio of same to different author pairs of all datasets is 1:1.

\section{Results}

\label{sec:results}

\paragraph{Metrics} We evaluate \rsp against both \gtv and neural models on the receiver-operating curve area-under-curve (AUC), which represents a model's performance across all thresholds. It calculates the true positive rate (TPR) and false positive rate (FPR) at every threshold, and graphs TPR over FPR. We use AUC as it is threshold-independent and the data we use is balanced, providing a direct comparison of the various systems. 
%\pegah{Maybe bring this explanation of auc to the experimental setup?} 

% \includegraphics
% We present the AUC curves of the three methods we evaluate on all four datasets in Figure \ref{fig:auc}.

% \noindent\textbf{Reddit}
% \begin{figure}[htbp]
%     \centering
%     \includegraphics[width=0.45\textwidth]{AUC-curves/reddit_auc.png}
%     \caption{ROC AUC Curve for \gtv, LUAR, \rsp on Reddit Dataset. We observe a 17 point increase in performance from \gtv using \rsp.}
%     \label{fig:reddit_auc}
% \end{figure}

% \noindent\textbf{Amazon}
% \begin{figure}[htbp]
%     \centering
%     \includegraphics[width=0.45\textwidth]{AUC-curves/amazon_auc.png}
%     \caption{ROC AUC Curve for \gtv, LUAR, \rsp on Amazon Dataset. We observe a 25 point increase in performance from \gtv using \rsp.}
%     \label{fig:reddit_auc}
% \end{figure}

% \noindent\textbf{Fanfiction}
% \begin{figure}[htbp]
%     \centering
%     \includegraphics[width=0.45\textwidth]{AUC-curves/fanfiction_auc.png}
%     \caption{ROC AUC Curve for \gtv, LUAR, \rsp on Fanfiction Dataset. We observe a 18 point increase in performance from \gtv using \rsp.}
%     \label{fig:reddit_auc}
% \end{figure}

% \noindent\textbf{Pikabu}
% \begin{figure}[htbp]
%     \centering
%     \includegraphics[width=0.45\textwidth]{AUC-curves/pikabu_auc.png}
%     \caption{ROC AUC Curve for \gtv, LUAR, \rsp on Pikabu Dataset. We observe a 11 point increase in performance from \gtv using \rsp, and notably, a 2 point increase in performance from LUAR using \rsp.}
%     \label{fig:reddit_auc}
% \end{figure}

\subsection{System Evaluation} We show the performance of our \rsp system in Table \ref{tab:full-results}. The results of \gtv alone are given in the first row. We then show the results of testing seven different neural models on three English corpora, and one neural model on a Russian corpus. For each neural model and dataset, we present two results, the first being the neural baseline, and the second being our \rsp system performance. For all combinations of datasets and models, we see that the uninterpretable neural system and the partially interpretable \rsp system perform approximately similarly, with no clear pattern emerging with respect to the neural system and/or to the corpus.
We see that \gtv performs consistently worse than any neural system or any of our \rsp systems, but above the random baseline (0.5).

We see that \luar outperforms all other neural models for Reddit and Amazon, and performs competitively for Fanfiction. This is expected because \luar is trained for authorship attribution. Furthermore, \luar is particularly good for Reddit, which is also expected, as \luar is trained on Reddit exclusively. Therefore, we now concentrate on \luar.

The performance for \rsp using \luar and the uninterpretable \luar neural baseline  are nearly identical in AUC for each dataset, with \rsp performing slightly worse for Reddit, and slightly better Pikabu. However, we observe a big increase in performance compared to using \gtv alone, with the biggest improvement being an increase of 19 points on the Amazon dataset.\footnote{We perform significance tests (paired bootstrap on AUC, two-sided) to compare \rsp to Gram2Vec, and for all four corpora this difference is significant with p < 0.001.}  

\subsection{Architecture Ablation}
We experiment with two alternative residual architectures: (i) a simpler version that passes only the neural embeddings into the regression head, leveraging the representation power of language models like RoBERTa for sequence classification through the [CLS] token, and (ii) a variant that directly appends the interpretable feature vectors to the neural embeddings before passing into the regression head to predict the residual. The latter incorporates signal directly from the interpretable system into the training of \rsp. We provide an ablation study comparing these two variants with our final version in Table \ref{tab:ablation}.

\begin{table}[h]
\setlength\tabcolsep{4pt}
\centering
\small
% \label{rsp_results}
\begin{tabular}{lcccc}  % No vertical rules
\toprule
\textbf{Data}  & \textbf{RS} & \textbf{Only-Neural} &  \textbf{Appended} \\
\midrule
Reddit      & 0.80 & 0.73 & 0.73 \\
Amazon      & 0.90 & 0.84 & 0.85 \\
Fanfiction  & 0.87 & 0.74 & 0.77 \\
\bottomrule
\end{tabular}
\caption{An ablation study of the different variants of our system. This table is a comparison of our final \textbf{residualized similarity (RS)} system, which uses an attention layer to combine the interpretable feature vectors with the neural embeddings, compared to an initial version that passed the neural embeddings directly into the regression head (\textbf{Only-Neural}), and another version that concatenated the neural embeddings to the interpretable features directly (\textbf{Appended}).}  
% We evaluate on Residualized Similarity AUC (\textbf{RS-AUC}) and Contrastive Loss Fine-Tuned AUC (\textbf{CL-AUC}). \textsuperscript{$\dagger$} denotes a Russian multilingual dataset.}
\label{tab:ablation}
\end{table}

\begin{table*}[h]
\setlength\tabcolsep{4pt}
\centering
\small
\begin{tabular}{lccccccc}  % No vertical rules
\toprule
\textbf{Data}  & \textbf{\gtv} & \textbf{\gtv-RS} &  \textbf{ELFEN} & \textbf{ELFEN-RS} & \textbf{Combined} & \textbf{Combined-RS}  \\
\midrule
Reddit      & 0.63 & 0.80 & 0.59 & 0.80 & 0.63 & 0.80\\
Amazon      & 0.71 & 0.90 & 0.71 & 0.90 & 0.74 & 0.88 \\
Fanfiction  & 0.69 & 0.87 & 0.66 & 0.88 & 0.70 & 0.86 \\
\bottomrule
\end{tabular}
\caption{The results of a \textbf{residualized similarity (RS)} system with \gtv, ELFEN, and the concatenation of both as the interpretable system. While the concatenation of the two tends to perform a bit better than either feature set alone, the RS system trained on top of the concatenation performs worse than either RS system trained on a single interpretable system.}
\label{tab:elfen}
\end{table*} 

\begin{figure}[h]
    \centering
    \includegraphics[width=\columnwidth]{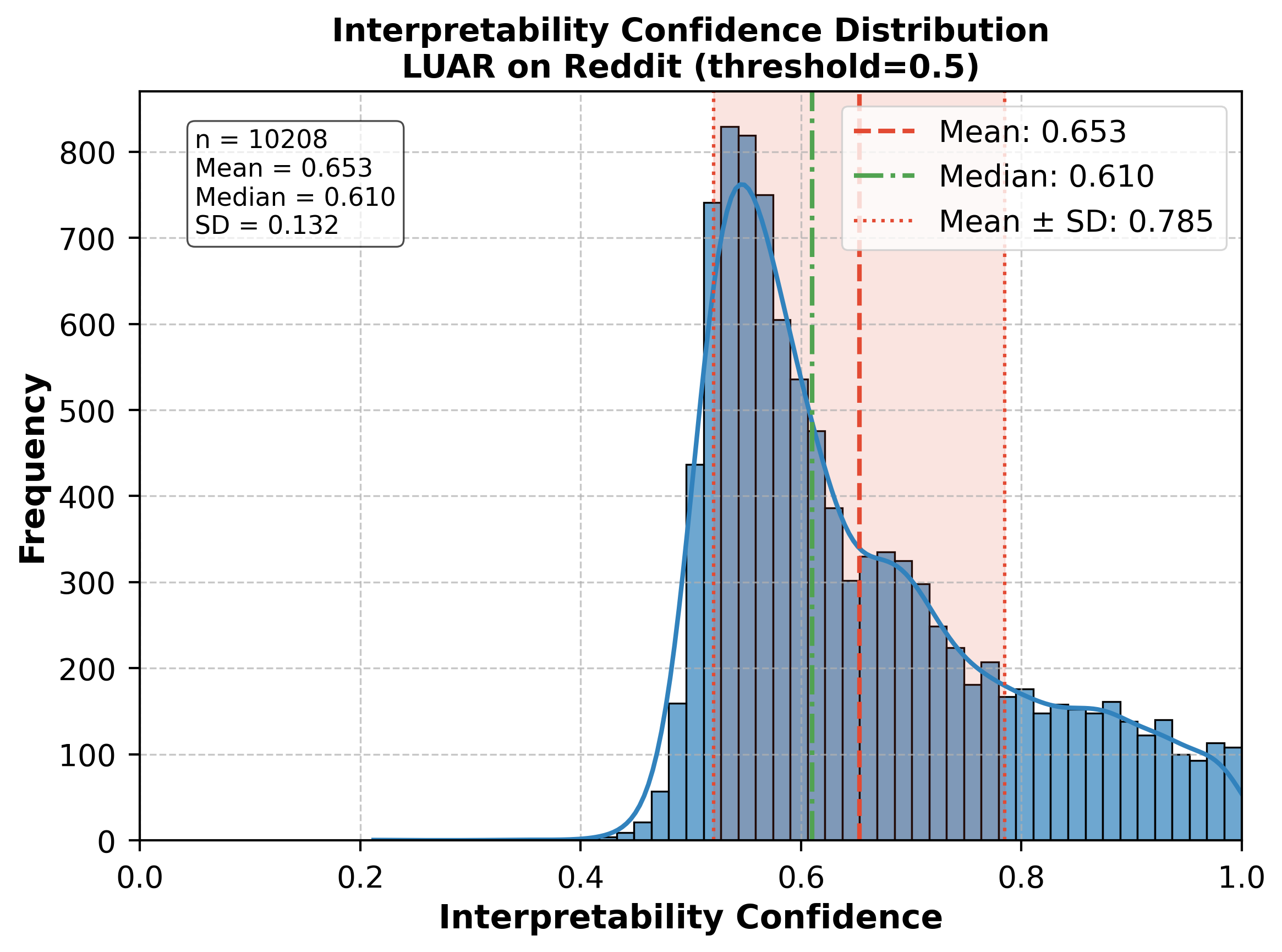}
    \caption{The distribution of interpretability confidence scores in the predictions using \rspfull with LUAR on the Reddit dataset.}
    \vspace{-5pt}
    \label{fig:ic_distribution}
\end{figure}

\subsection{Analysis of Interpretability Confidence}
We plot the distribution of interpretability confidence scores in the predictions using \rsp with LUAR on the Reddit dataset shown in figure~\ref{fig:ic_distribution}. Notably, we observe in most cases, we have an \ic of greater than 0.5, and a mean of .65.  This suggests that our \rsp system retains a useful amount of interpretability from the \gtv system while increasing performance.

% \noindent{\textbf{Correlation Study between \ic Threshold and Accuracy}} We consider the relationship between \ic and the accuracy of the predictions that remain above the threshold.\owen{You need to explain in more detail here what the data is -- 500 test set pairs, each with an IC, then you can threshold the IC, then you get fewer and fewer pairs, etc} We remove points with \ic less than 0.05 and greater than 0.95, and observe a Pearson r of 0.994 (p = 4.5e-17), indicating a strong correlation between the \ic threshold and the accuracy of the remaining points. Details are shown in Figure \ref{fig:ic_threshold_correlation}. 

% \noindent{\textbf{Analysis of Prediction Flipping}} We look at the contingency table of the predictions for the Reddit dataset, and measure the Chi-square statistic and odds ratio. 

% \begin{table}[htbp]
%     \centering
%     \begin{tabular}{lcc}
%         \toprule
%         \textbf{Prediction Flipped} & \textbf{Incorrect} & \textbf{Correct} \\
%         \midrule
%         False & 785 & 3,029 \\
%         True & 2,284 & 4,110 \\
%         \bottomrule
%     \end{tabular}
%     \caption{Contingency Table: Prediction Flips vs Correctness for \rsp fine-tuned on the Reddit Dataset. A chi-square test of independence revealed a significant relationship between prediction flips and correctness ($\chi^2(1) = 259.69$, $p < .001$). The odds ratio of 2.14 indicates that non-flipped predictions were more than twice as likely to be correct compared to flipped predictions.}
%     \label{tab:contingency}
% \end{table}

\section{Robustness of the Technique}
To show that our technique is not dependent on any single interpretable base system, we try another feature extractor in place of \gtv. We turn our attention towards Efficient Linguistic Feature Extraction for Natural Language Datasets, or \elfen{}~\citep{maurer-2025-elfen}. While some features overlap, \elfen captures several unique categories of features from \gtv, and as such, provides a good alternative interpretable system to use in our \rsp framework. \elfen feature areas include dependency, emotion, entities, information, lexical richness, morphological, part-of-speech, readability, semantic, and surface. We provide brief explanations of the different features in Appendix~\ref{sec:appendix_elfen}.

We perform experiments using \elfen in place of \gtv as the interpretable system, as well as the concatenation of the two interpretable feature sets as the interpretable system. We perform these experiments using LUAR as the neural model as it is generally the strongest performing neural model of our primary experiments. We find that on its own, \elfen is a little less performant or on par with \gtv, while concatenating the two feature sets tends to improve a bit on either. However, notably, the \rsp model performs about the same regardless of the interpretable system that it is trained on. We consider this to be a success of the \rspfull technique as it is able to boost the performance of two separate interpretable systems.\footnote{We perform significance tests (paired bootstrap on AUC, two-sided) to compare \gtv-RS to \gtv alone, ELFEN-RS to ELFEN alone, and Combined-RS to Combined alone. For all three corpora and all three pairs of systems, this difference is significant with p < 0.001.}

% \section{Meta-level Analysis of \gtv Features}
\section{Case Study of Two Pairs of Documents}
\label{sec:example}

\begin{figure}[ht!]
    \begin{tcolorbox}[
        colback=yellow!15!brown!10!white,    
        colframe=brown!20!white!80, 
        arc=5mm,
        width=\columnwidth,
        before upper={\parindent15pt},
    ]
    \small
    \noindent\textbf{Example Pair 1: Different Author}
    \vspace{0.5em}
    
    \noindent\textbf{Document 1:}\\
    Whirling like  \highlight{a} scythe, the saber sliced her upper torso, putting an end to the \highlight{vengeful Sith}. Dropping to her knees again, Jameh crawled to her \highlight{fallen Master}, cradling him in her arms.  \highlight{A} new darkness grew in her heart now, one like a cold, lonely mist. 
    %Her Master was dying. Just then, footsteps came down the cave passage and Pilae, Obi-Wan, and Anakin entered the grotto just in time to be too late. 
    Pilae, Obi-Wan, and Anakin stood nearby, dismayed at the sight that met their eyes:  \highlight{a}  dismembered \highlight{former Senator}, \highlight{a} shorn and \highlight{wounded Padawan}, and  \highlight{a} Jedi Master on the verge of death. " Master, please, you can"t leave me. I need you; I"m not ready!"
    
    \medskip
    \noindent\textbf{Document 2:}\\
    %As the Clan speculated why the rats weren"t attacking, Redfur walked through the camp entrance tentatively, leaving Sootcloud and Brightnose at their original position at one side of the entrance. 
    He scanned the field beyond and was dumbfounded \highlight{when} he didn"t see any rats. 
    %As he walked further out with more confidence, he tasted the air and searched for their distinctive scent. Suddenly, with a loud squeak, several of the rats surged forward out of nowhere, or so Redfur thought, and attacked him. He yowled in surprise as some of the rats managed to climb up his leg and cling to his red brown fur, leaving scratches and bites along the way. 
    He pelted back through the entrance and into the clearing. The Clan had been alerted by Redfur"s yowl of surprise, so they had stopped chatting and lowered their bodies into a crouch, getting ready for the rats. But \highlight{when} they saw the four rats clinging to Redfur"s fur, they hissed in astonishment at the size of them.
    \vspace{0.5em}
    
    \noindent\textbf{\gtv Cosine Similarity}: 0.09,
    \noindent\textbf{\rsp Predicted Residual}: 0.29,
    \noindent\textbf{Final Score}: 0.38 \\
    \noindent\textbf{Interpretability Confidence}: 0.76
    \noindent\textbf{Flipped}: False
    \vspace{0.5em}
    \hrule
    \vspace{0.5em}
    
    \noindent\textbf{ Example Pair 2: Same Author}
    \vspace{0.5em}
    
    \noindent\textbf{Document 1:}\\
    GET \highlight{UP!} School time!" Sora called from the door. " I"m \highlight{up!}" he hollered back before throwing the cover"s off him. It"s been a week. A week since Roxas started hearing that voice. Throughout that time he had figured out that \highlight{it was connected} to the mirror he had gotten at the same time. "
    
    \medskip
    \noindent\textbf{Document 2:}\\
    \highlight{It was passed down} through generations to keep him in the glass." At this he closed the book and plopped on the bed. " What about the rhyme?" Demyx stroked his chin in a pondering position. " \highlight{It was created} to scare children from letting him out. Though the ending part. "" A curse to never be free \highlight{of.} Until this demon admits love" Is exactly what it says.

    \vspace{0.5em}
    \noindent\textbf{\gtv Cosine Similarity}: 0.20, 
    \noindent\textbf{\rsp Predicted Residual}: 0.82,
    \noindent\textbf{Final Score}: 1.02 \\
    \noindent\textbf{Interpretability Confidence}: 0.59
    \noindent\textbf{Flipped}: True
    \end{tcolorbox}
    \caption{Example Pairs for Case Study. Pair 1 is by two different authors, and Pair 2 is by the same author.}
    \label{fig:example_pairs}
    \vspace{-5pt}
\end{figure}
We present two cases to illustrate how \rsp can give a user insight while performing a specific authorship verification task. We present two pairs of documents, one of which is indeed from the same author, and one of which is not. Given the range of cosine similarity from -1 to 1, we set a threshold of 0.5, representing a moderately strong alignment in vector space and
% \owen{Why 0?  Range of cosine is -1 to 1.  I think we need to say that instead it gives a good result (but what did we tune on?)}, 
suggesting that documents need to be more similar than dissimilar to be considered by the same author.  We show how our approach can tell the user which \gtv features were used in the determination, and to what extent they determined the confidence of the prediction. Since \gtv contains over 600 features, we define a criterion to select features to present to the user, depending on whether a pair of documents are predicted to be by the same or different authors. When a pair of documents is predicted to be written by the same author, we want to maximize the absolute values of the feature values (features that distinguish these documents from the large set of background documents) while making sure the values are similar for both documents. When a pair of documents is predicted to be written by different authors, we simply find the largest magnitudes of differences in the feature values. Thus, for identifying features for same-author pairs, we use the following metrics for ordering features, where $val\_1$ represents the feature's score for document 1, and $val\_2$ represents the feature's score for document 2.: $|val\_1| + |val\_2| - |val\_1 - val\_2|$. For ordering features using different author pairs, we use $|val\_1 - val\_2|$.  We then choose the top $n$ features; in the examples below, we use $n=5$.
\begin{table}[h]
    \centering
    \small
    \begin{tabular}{@{}l@{\hspace{3pt}}r@{\hspace{3pt}}r@{\hspace{3pt}}r@{}}
    \toprule
    \textbf{Feature} & \textbf{Score} & \textbf{Doc 1} & \textbf{Doc 2} \\
    \midrule
    func\_words:further & 5.4 & -0.1 & 5.3 \\
    pos\_bigrams:ADJ PROPN & 4.1 & 3.8 & -0.3 \\
    pos\_bigrams:PUNCT DET & 3.8 & 3.4 & -0.4 \\
    %func\_words:through & 3.6 & -0.3 & 3.3 \\
    %pos\_bigrams:PART ADJ & 3.3 & 3.1 & -0.2 \\
    morph\_tags:Definite=Ind & 2.8 & 2.4 & -0.4 \\
    func\_words:when & 2.6 & -0.4 & 2.2 \\
    \midrule
    pos\_bigrams:PREP PUNCT & 6.1 & 4.1 & 3.0 \\
    passive sentence & 5.4 & 2.7 & 4.6 \\
    dep\_labels:nsubjpass & 4.4 & 2.2 & 3.8 \\
    pos\_bigrams:PREP VERB & 4.3 & 2.9 & 2.1 \\
    %dep\_labels:auxpass & 3.7 & 1.8 & 3.3 \\
    punctuation:, & 3.4 & -1.7 & -1.7 \\
    % func\_words:they & 2.9 & -0.4 & 2.5 \\
    % pos\_bigrams:PROPN PUNCT & 2.9 & 2.1 & -0.8 \\
    % 
    % pos\_bigrams:PUNCT NUM & 2.6 & 2.5 & -0.1 \\
    % func\_words:when & 2.6 & -0.4 & 2.2 \\
    \bottomrule
    \end{tabular}
    \caption{Top half: The feature scores comparison between the Example 1 document pair by different authors. Bottom half: The feature scores comparison between the Example 2 document pair by the same author.}
    \label{tab:author_scores}
\end{table}

\subsection{Example 1: Different Author Pair}
\label{sec:example_1}

In the first example in Figure~\ref{fig:example_pairs}, 
%based on a threshold of 0.5, 
both \gtv and \rsp predict that these two documents are written by different authors: 
%the gold label for different authors is -1, and we see that in this case, 
the \gtv similarity is $0.09 < 0.5$ and \rsp's is $0.09 + 0.29 = 0.38 < 0.5$, and thus that the label is not flipped. Since the prediction is ``different author", we use the formula defined in Section~\ref{sec:interpretability} to calculate the \ic as: $$\frac{1-0.09}{1-0.09+0.29} = 0.76$$ giving us a high confidence in the interpretability. In the top half of Table~\ref{tab:author_scores}, we show the top 5 features and their values that were identified using the different author pair metric: $|val\_1 - val\_2|$. We calculate this score for every feature in document 1 and document 2, and sort the top 5 features in descending order. These represent the 5 most differing features in the pair of documents.  Looking at the features, we first note several function words which can be found in document 2 but not in document 1; for example, document 2 uses {\em when} twice in a fairly short text, while document 1 does not use it at all.  In contrast, document 1 uses several part-of-speech (POS) bigrams far more frequently than the background corpus, while document 2's distribution of POS bigrams is more standard.  A striking example is the bigram adjective-proper noun, which is unusual in general but very frequent in document 1 ({\em vengeful Sith}, {\em fallen Master}, {\em former Senator}, {\em wounded Padawan}).  Finally, we note the high frequency of the indefinite article in document 1: {\em a scythe}, {\em a new darkness}, {\em a cold, lonely mist}, {\em a dismembered former senator}, {\em a shorn and wounded Padawan}, {\em a Jedi Master}.  
These indefinite noun phrases provide a sense of change (indefinites introduce new discourse objects).
%; in the case of the last three, the author takes on the perspective of three characters. 
Document 2, in contrast, has few indefinites, and the narration centers on entities known to the readers and the characters in the story.

\subsection{Example 2: Same Author Pair}

%In this case, based on a threshold of 0.5, we observe that 
\gtv predicts that the two documents are written by different authors, getting the prediction wrong on its own.  But overall, \rsp predicts correctly that these two documents are written by the same author. 
%The gold label for the same author is 1, so \gtv gets the prediction wrong. However, \rsp predicts the similarity residual and the final score is right at the gold label for the same author. 
Since the prediction is ``same author", we use the formula defined in Section~\ref{sec:interpretability} to calculate the \ic as: $$\frac{1+0.20}{1+0.20+0.82} = 0.59$$ giving us a moderately high confidence in the interpretability.
Even though the label was flipped from \gtv to \rsp in this case, we observe that there are still features that are similar between the two documents, which we can use in explanation, since they in fact contributed to the final prediction. When identifying similar features in two documents, we use the metric $|val\_1| + |val\_2| - |val\_1 - val\_2|$ and take the top 5 features in descending order, shown in the bottom half of Table~\ref{tab:author_scores}.
Thus, these are features which occur in both documents either much more or much less frequently than on average across a background corpus.  One example is the bigram preposition-punctuation.  In both texts, we find examples: {\em UP!}, {\em up!} (in document 1), {\em out.}, {\em of.} (in document 2).  
%A preposition at the end of a clause is often discouraged in formal written English.   
The two documents also use passive voice clauses more frequently than on average:
%(passive voice is generally rare in written English): 
{\em it was connected} (document 1), {\em it was passed down}, {\em it was created} (document 2).  Both of these linguistic features are relatively rare in standard written English.
The two documents share a negative value for the punctuation mark comma.  Indeed, neither text contains a comma, which in general is a very common punctuation mark.

% \begin{table}[htbp]
%     \centering
%     \small
%     \begin{tabular}{@{}l@{\hspace{3pt}}r@{\hspace{3pt}}r@{\hspace{3pt}}r@{}}
%     \toprule
%     \textbf{Feature} & \textbf{Score} & \textbf{Doc 1} & \textbf{Doc 2} \\
%     \midrule
%     pos\_bigrams:PREP PUNCT & 6.1 & 4.1 & 3.0 \\
%     passive sentence & 5.4 & 2.7 & 4.6 \\
%     dep\_labels:nsubjpass & 4.4 & 2.2 & 3.8 \\
%     pos\_bigrams:PREP VERB & 4.3 & 2.9 & 2.1 \\
%     dep\_labels:auxpass & 3.7 & 1.8 & 3.3 \\
%     % func\_words:from & 3.5 & 2.3 & 1.8 \\
%     % punctuation:, & 3.4 & -1.7 & -1.7 \\
%     % morph\_tags:PunctType=Comm & 3.3 & -1.6 & -1.6 \\
%     % pos\_bigrams:DET NOUN & 3.2 & 2.5 & 1.6 \\
%     % func\_words:the & 2.9 & 1.5 & 1.5 \\
%     \bottomrule
%     \end{tabular}
%     \caption{Feature scores comparison between Example 2 document pair by the same author.}
%     \label{tab:same_author_scores}
% \end{table}

\label{sec:example_2}

We note that this paper does not propose an end-to-end explainable system.  Instead, we have shown how our \rsp system can identify measurable features which it actually used in determining its finding (faithfulness), and it can quantify to what extent these features explain why the system came to its result.

\section{Conclusion}
% \pegah{since the paper is now very focused on authorship verification we should probably update this part accordingly}
We introduce \rspfull, a method of improving the performance of an interpretable feature set by training a language model to predict the residual, or difference, between the similarity output from an interpretable system and the ground truth. We apply this technique to the task of authorship verification, where interpretability is of the utmost importance. Using \rspfull, we are able to achieve state-of-the-art performance on the task of authorship verification while maintaining a quantifiable degree of interpretability. 

To measure interpretability, we introduce the \textbf{interpretability confidence}, a measure of how interpretable a prediction from our system is. We then do a case study to observe how using \rsp, we are able to correct a prediction that was initially incorrect from an interpretable system. In both the case where the prediction was corrected and the case where the prediction from the interpretable system and \rsp agreed, we show that there is meaningful interpretability in the features, as well as the ability to trace such features back to the text from which they were extracted.

We believe this approach to be a promising direction for developing more interpretable and effective NLP systems, bridging the gap between neural methods and interpretable linguistic features while allowing for faithfully explainable systems.

\section*{Limitations}
We present preliminary results on \rspfull (\rsp), a novel method of supplementing systems using interpretable linguistic features with a neural network to improve their performance while maintaining interpretability. In order to get these results, we use a relatively small subset of data from the original datasets we chose. While we choose a variety of datasets, our experiments are by no means conclusive. 

An explainable system built on top of our system would require, in addition, two types of decisions: how do we choose how many and which features to present to the user, and exactly how should the interface look?  These are, at base, human-computer interface (HCI) issues: explanations are always for a particular type of user, and need to be tailored to that user.  If, for example, our target audience is forensic linguists, then we can assume that they know the meaning of linguistic features and are willing to get to know a more complex interface (which, for example, may allow them to drill down, or to include or exclude certain types of linguistic features).  If, on the other hand, the target audience is crowdsourced workers (because we are evaluating a paper for a submission to an NLP conference, for example), then of course we cannot assume the users will know the meaning of our features, nor that they will take the time to get to know the capabilities of a more complex interface.  We leave this HCI work to a future publication.  

\section*{Ethics Statement}
The underlying datasets we use are publicly available and are anonymized. Our work improves the interpretability of authorship verification models, allowing for more transparency and easier detection of potential biases and errors in the model.

\section*{Acknowledgements}
This material is based upon work supported in part by the National Science Foundation (NSF) under No. 2125295 (NRT-HDR: Detecting and Addressing Bias in Data, Humans, and Institutions); as well as by the Intelligence Advanced Research Projects Activity (IARPA) under the HIATUS program (contract 2022-22072200005). Any opinions, findings, and conclusions or recommendations expressed in this material are those of the author(s) and do not necessarily reflect the views of the NSF, DARPA, or IARPA. 

We thank both the Institute for Advanced Computational Science and the Institute for AI-Driven Discovery and Innovation at Stony Brook for access to the computing resources needed for this work. These resources were made possible by NSF grant No. 1531492 (SeaWulf HPC cluster
maintained by Research Computing and Cyberinfrastructure) and NSF grant No. 1919752 (Major Research Infrastructure program), respectively.

We thank our ARR reviewers, whose comments
have contributed to improving the paper.

\bibliography{anthology,custom}
% \bibliography{custom}
\newpage
\appendix

\section{Training Details}
\label{sec:appendix_training}
We experiment with a variety of strategies to decrease training times and GPU memory requirements. All our experiments take place on a server with four 48GB A6000 GPUs. Using the following strategies, our largest model, with approximately 360 million parameters, takes about 5 hours to train. The fastest training time we observed was around 1 hour for our smaller models, which have approximately 150 million parameters. We optimize the model using AdamW \citep{loshchilov2017decoupled} with a learning rate of 5e-5, a standard value for fine-tuning pre-trained language models. We train for a maximum of 10 epochs with early stopping based on validation loss to avoid overfitting. 
With respect to hyperparameters, we manually tune them during the training of \rsp. We use these hyperparameters in the rest of our experiments.

We experiment with the use of LoRA \citep{hu2021lora}, reducing the number of trainable parameters and lowering memory requirements. Somewhat surprisingly, in our initial experiments, fine-tuning RoBERTa for binary classification and for our residual prediction model, performance without LoRA was far lower than performance using LoRA. We hypothesize that LoRA could be acting as a regularizer in this case. We use this to inform our decision to use LoRA in all other experiments in this paper.

\paragraph{Neural Model Contrastive Loss Fine-Tuned Baseline} We fine-tune the previously chosen neural models in a Siamese network using a contrastive loss function as our training objective. The architecture for this was heavily inspired by SBERT \citep{reimers-gurevych-2019-sentence}. We replace SBERT with \luar or \luarru, and use the pooler output to obtain the embedding for the documents.

\paragraph{Residualized Similarity Details} As \rsp is a regression model, we use mean-squared error loss as our training objective, and train over 10 epochs. We utilize early stopping to avoid over-fitting. We add a regression head with multiple dense layers using ReLU activations and dropout for regularization. We then ensure the output is between -1 and 1 by using a tanh activation.
\label{sec:training-details}

\section{Cross-Domain Experiments}
\label{sec:appendix_cross_domain}
We evaluate \rsp models trained on one domain on the other domains, similar to the cross-domain experiments in the LUAR paper \citep{rivera-soto-etal-2021-learning}. In addition, we evaluate a \rsp model trained on the English Reddit data on the Russian Pikabu data to observe if there is any cross-lingual transfer.

\begin{table}[h]
\centering
\begin{tabular}{lcccc}
\toprule
\textbf{Trained On} & \textbf{Dataset} & \textbf{G2V} & \textbf{RSP} \\
\midrule
\multirow{3}{*}{Reddit} & Reddit & 0.63 & 0.80 \\
& Amazon & 0.71 & 0.81 \\
& Fanfiction & 0.69 & 0.71 \\
\midrule
\multirow{3}{*}{Amazon} & Reddit & 0.63 & 0.71 \\
& Amazon & 0.71 & 0.90 \\
& Fanfiction & 0.69 & 0.72 \\
\midrule
\multirow{3}{*}{Fanfiction} & Reddit & 0.63 & 0.67 \\
& Amazon & 0.71 & 0.75 \\
& Fanfiction & 0.69 & 0.87 \\
\midrule
Reddit & Pikabu & 0.65 & 0.62 \\
\bottomrule
\end{tabular}
\caption{Cross-domain experiments. Unsurprisingly, models trained on one domain tend to do a bit worse when evaluating on the other two domains. However, the performance in all cases still beat using just \gtv. We also witness no cross-lingual transfer, suggesting a model trained on English data would not be able to help in evaluation performance on Russian data.}
\label{tab:performance_metrics}
\end{table}

\section{\elfen Feature Categories}
\label{sec:appendix_elfen}
Efficient Linguistic Feature Extraction for Natural Language Datasets (\elfen) \citep{maurer-2025-elfen} is an interpretable feature extractor. We provide brief explanations of the different feature types; the full feature list can be found at the github\footnote{https://github.com/mmmaurer/elfen/blob/main/features.md}.

\begin{itemize}
    \item Dependency: dependency features are based on the dependency tree of the text data, with features such as tree width, tree depth, and more.
    \item Emotion: emotion features include valence, arousal, dominance dimensions, Plutchik emotions, the sentiment and emotion intensity of text data. Calculating these features uses the NRC VAD Lexicon \citep{mohammad2013crowdsourcing}, the NRC Emotion Intensity Lexicon \citep{mohammad-2018-obtaining}, and the NRC Sentiment Lexicon \citep{mohammad-2018-word}.
    \item Entities: entity features include named entity-related features such as 'PERSON', 'MONEY', 'PRODUCT', 'TIME', and 'PERCENT'.
    \item Information: this area includes information-theoretic metrics, in particular, compressibility and Shannon Entropy.
    \item Lexical Richness: this area contains various lexical richness metrics such as lemma/token ratio, type/token ratio, and root type/token ratio.
    \item Morphological: this area captures the counts of tokens with a specific type of morphological tag such as "VerbForm" or "Number".
    \item Part-of-Speech: this area includes various part-of-speech (POS) related features such as POS variability and number of tokens per POS tag.
    \item Readability: this area includes different readability/complexity scores such as number of syllables and number of mono/poly-syllables.
    \item Semantic: this area includes different types of semantic features such as number of hedge words and ratio of hedge words.
    \item Surface: this area includes ``surface-level" features such as raw sequence length, number of tokens, and number of sentences. 
\end{itemize}
\end{document}